# EVOLUTIONARY DEMOGRAPHIC ALGORITHMS

**MARCO AR ERRA, PEDRO MM MITRA, AGOSTINHO C ROSA**

Laseeb-ISR, DEEC-IST
Av. Rovisco País, 1, Torre Norte 6.21.1049, Portugal
mare@mega.ist.utl.pt
pmmmi@mega.ist.utl.pt
acrosa@laseeb.org

## ABSTRACT

Most of the problems in genetic algorithms are very complex and demand a large amount of resources that current technology can not offer.

Our purpose was to develop a Java-JINI distributed library that implements Genetic Algorithms with sub-populations (coarse grain) and a graphical interface in order to configure and follow the evolution of the search. The sub-populations are simulated/evaluated in personal computers connected trough a network, keeping in mind different models of sub-populations, migration policies and network topologies. We show that this model delays the convergence of the population keeping a higher level of genetic diversity and allows a much greater number of evaluations since they are distributed among several computers compared with the traditional Genetic Algorithms.

## INTRODUCTION

The distributed implementation of the evolutionary genetic algorithm has two major advantages: the increase of the speed of execution from the direct parallelization and the diversity of subpopulations effect that tends to avoid the premature convergence effect (this last one, a problem with a difficult solution in almost all evolutionary algorithms).

The Laboratório de Sistemas Evolutivos e Engenharia Biomedica (Laseeb), in Instituto Superior de Robotica (ISR), where the project was developed already had a library that implements the most common variants of evolutionary algorithms and a master-slave distributed model called JDEAL (Java Distributed Evolutionary Algorithm Library). Our project was built using this library, extending and complementing it with new functionalities.

A too fast convergence would lead quickly to a solution but probably neither a good nor an efficient one, therefore the need of keeping the diversity of individuals in a population.

## DISTRIBUTED EVOLUTIONARY ALGORITHM

To solve the problem we present a solution based on islands. This solution allows the execution of several algorithms running in parallel in different machines or locally, but completely independent form each other, so that an island (algorithm) may exchange individuals with other islands in order to maintain diversity. These islands communicate between them through a network. The role of each island, to who can or can not communicate is dependent from the predefined topology.

When, how and who may or can migrate define a migration policy, which can influence positively or negatively the search objective.

Distributed evolutionary algorithms are normally used to solve complex problems where a normal computation power is not enough to cover the entire domain of possible states in order to achieve the best solution.

Although factors such as speed and search space size are important, there is no doubt that the risk of the algorithm falling into a local maximum is a factor to avoid. In such situation the search doesn't evolve in a larger search space and the solution found may not be the optimum solution. If there is only a local maximum such problem is not so evident, as can been seen in figure 1, but if there is more than one and an individual fall in that maximum it will influence the rest of the population and the all search will converge to that value, when another and better one exists, as can be seen in figure 2. This problem is what the distributed approach seeks to avoid.

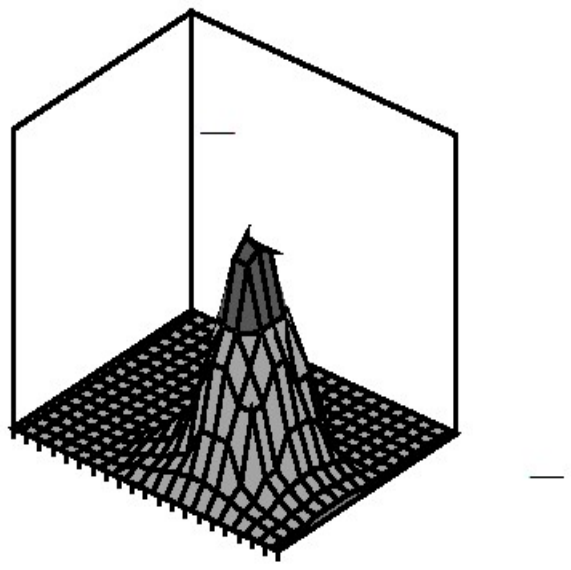

**Figure 1. Function with only one local maximum**

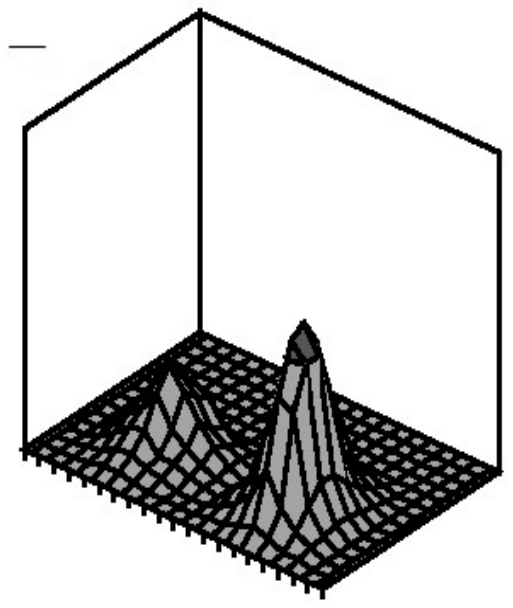

**Figure 2. Function with more than one local maximum**

The distributed genetic evolutionary algorithms can be divided in two categories: fine grain and coarse grain.

The fine grain distributed evolutionary algorithms can be represented in two ways. As a single population where each chromosome can only interact with its neighbors, or as sub-populations overlapped, where the interactions between the chromosomes are limited to the neighbor sub-populations.

The coarse grain distributed evolutionary algorithms, also known as the islands model, are a little more complex.

In this case the initial population is divided in sub-populations, and each sub-population evolves isolated from the others and occasionally exchanges chromosomes with the others. In this kind of problems it is usual to designate each sub-population as an island and the exchange as a migration between them.

The parallel implementation of the migration model shows not only a speed increase in the computation time, but also a reduction in the number of evaluation functions to apply, when compared to unique population algorithms. Therefore, even for a single processor computer, using this kind of parallel algorithm in a serial way (pseudo-parallel), better results can be achieved.

The selection of the individuals to migrate can be done randomly or fitness based.

Many types of topologies can be structured and implemented, being the most simple and general topology a no restrictions topology (each island can migrate with every other island).

In order to delay the propagation of a highly fit chromosome other strategies can be structured, for example a ring topology, in which each island can only exchange chromosomes with its neighbors.

## JAVA-JINI NETWORK TECHNOLOGY

We used Java-Jini network technology as a solution to the challenges posed by distributed systems: complexity, security, manageability, and unpredictability of computer networks. Jini offers network plug and play, which means that any device can be easily connected to the network, its presence announced and the clients that wish may use the services available. The mobility that Jini offers for distributed tasks over dynamic networks, where a service may be available for different periods of time, is the most suitable for this kind of problems.

A Jini community needs a set of support services, such as RMID to allow remote method invocation, an HTTPD to simplify stub transfers form the service provider to the service client, a *LookupService* in wich all services must register and where we can find all services available in the community and references of how to contact them and a *JavaSpaces* service that acts as a network storage repository in witch common data can be writen.

## IMPLEMENTATION

The project was divided in four modules: the demographic evolutionary algorithm, the communication based on Jini, the migration policies and the graphical interface.

The demographic evolutionary algorithm model implements a coarse grain distributed genetic algorithm, the communication model based on Jini makes available all the mechanisms needed to communicate and exchange between the evolutionary islands, the migration policies define how the chromosome exchange is done and when, finally the graphical interface allows the configuration and the monitorization of the evolution.

The tasks of configure and execute a genetic algorithm are complex and computationally heavy, in order to minimize the impact of these two factors and simplifying then, they were divided. A role called *researcher* was created with the task of configuring and collecting information about the search. Another role, called *worker*, was developed with the task of executing the algorithm.

Both of them register their Jini services in the *LookupService* to allow the services to be invoked remotely. When a well-behaved Jini service starts, it will register on all *LookupService* found.

A migration is an exchange of individuals between islands (*workers*). This exchange is made through the *JavaSpace*. With these new individuals we intend to introduce a larger heterogeneity to the destination population and consequentially enlarge the search space avoiding local maximums.

The graphical interface module makes easier the configuration and the monitorization of the search. A graphic allow us to follow the diversity of the individuals in all islands and the evolution of the search.

These extensions to the JDEAL library are composed by a set of new libraries (and some corrections to the previous libraries) that introduced new functionalities. These new libraries (communication, graphics, topologies, migrations, reports…) can also be extended if needed. This project is not only an application, already functional, but also a library for future, different and possible improvements and reference.

## RESULTS

The tests were a resolution of a scheduling problem more specifically a jobshop problem with a 15x20 matrix (15 machines for 20 jobs, problem swv11.job).

We performed four different tests in order to show how the migrations in the search could benefit the fitness and the deviation.

We used three computers. Two Intel Celeron 850 MHz with 382MB of RAM running each a worker and a AMD 1300 MHz with 512MB of RAM running the researcher with two local workers.

The tests were performed ten times each in order to achieve a representative average result.

All the tests had the following common parameters:

A population of 100 chromosomes, through 1000 generations, using one point crossover with 0.95% of probability and invert integer mutator with 0.05% probability. The topology used was with no restrictions and with no elitism.

In test1 it was used migrations every 100 generations, in test2 migrations every 200 generations, in test3 migrations every 500 generations and finally in test4 with no migrations.

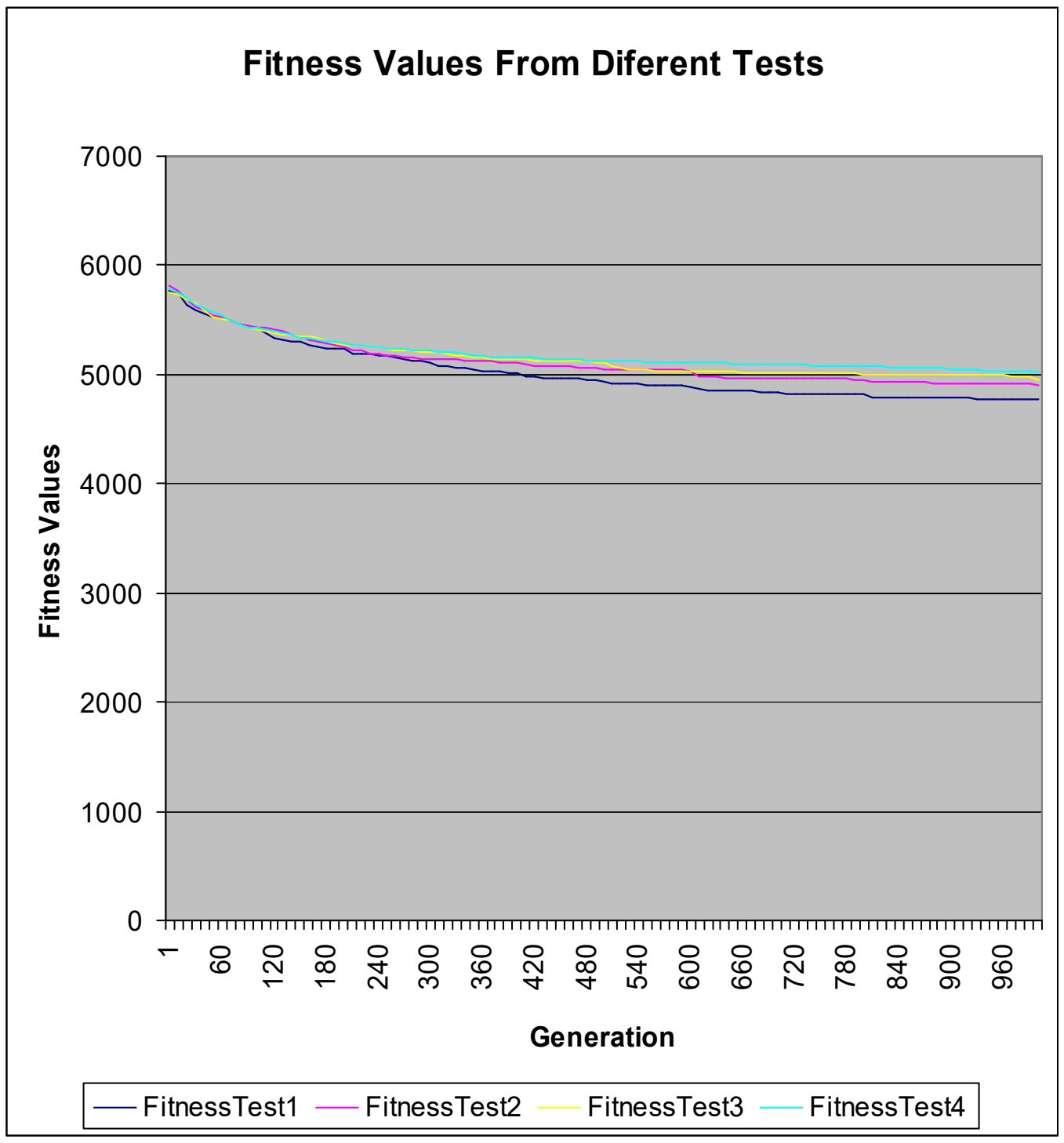

**Figure 3. Average Fitness**

In order to show a distributed genetic algorithm can run the same number of evaluations much faster than a normal genetic algorithm, we performed another set of tests.

Using the same computer (the AMD 1300 MHz described above) and running the same number of evaluations that four workers perform in a parallel execution another ten tests were performed in order to achieve a representative average result of the execution time.

The test1 was a population of 100 chromosomes, through 1000 generations, using one point crossover with 0.95% of probability and invert integer mutator with 0.05% probability. The topology used was with no restrictions and with no elitism and no migrations. An average number of 19450 evaluations was achieved.

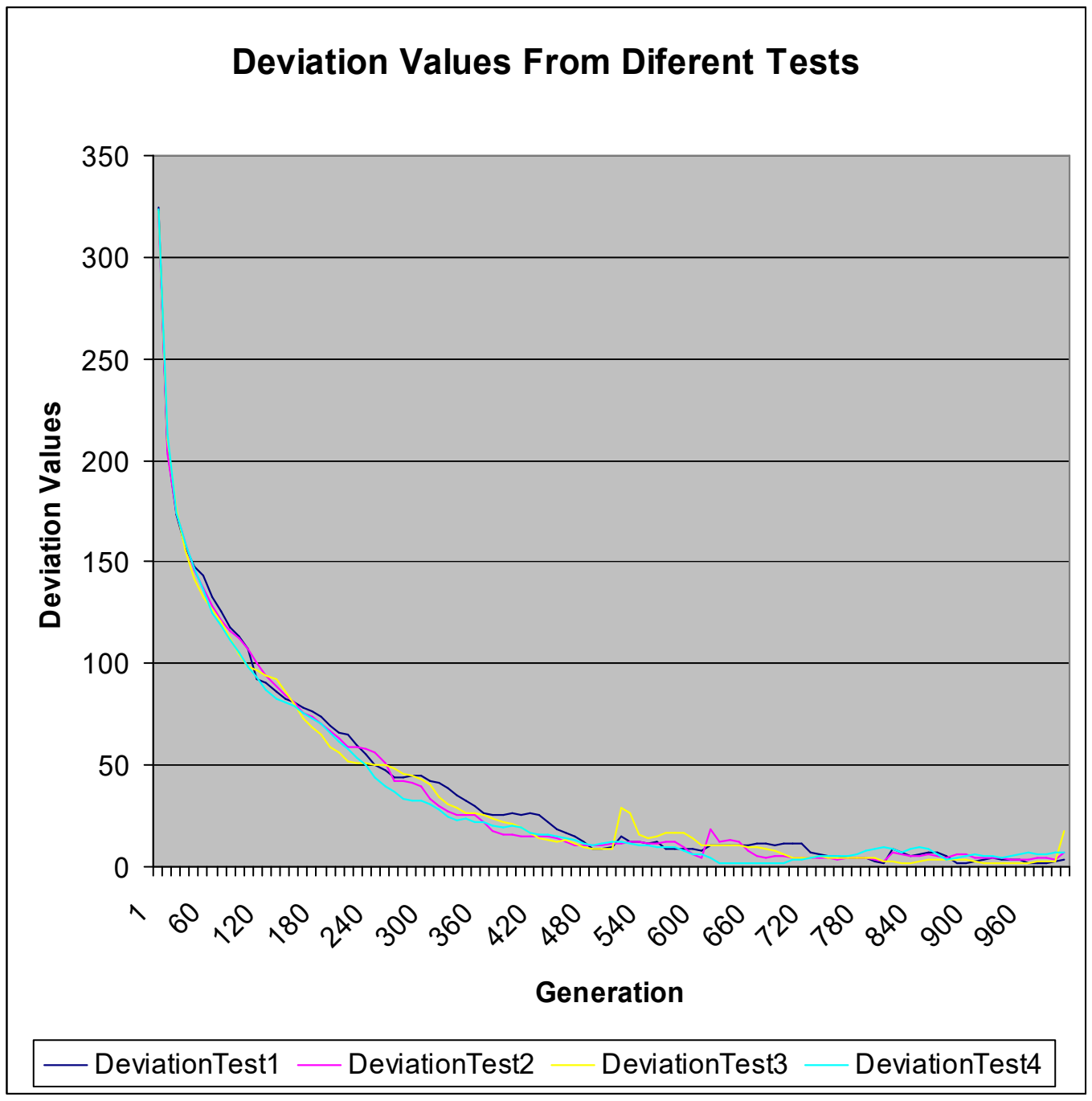

**Figure 4. Average Deviation**

The test2 was exactly the same but instead of 1000 generations as a conditional terminator an evaluation terminator with a value of 19450 evaluations on a single algorithm was used.

From the chart presented on Figure 3 we can see the evolution of the fitness value on the 4 sets of tests performed, with migrations every 100, 200 and 500 generations and a no migration search we obtained better results with the search were the migrations were made every 100 generations, but the higher heap in the deviation introduced by the arrival of new individuals was when the migrations were less frequent as we can see in the chart in figure 4 in witch we show the evolution of the deviation over the total number of generations.

The difference shown between the different algorithms in the fitness chart would tend to show higher values with the increase of the number of generations following the tendency denoted so far. If the study of this problem were to be more conclusive we would start by increasing the number of generations until we achieve convergence.

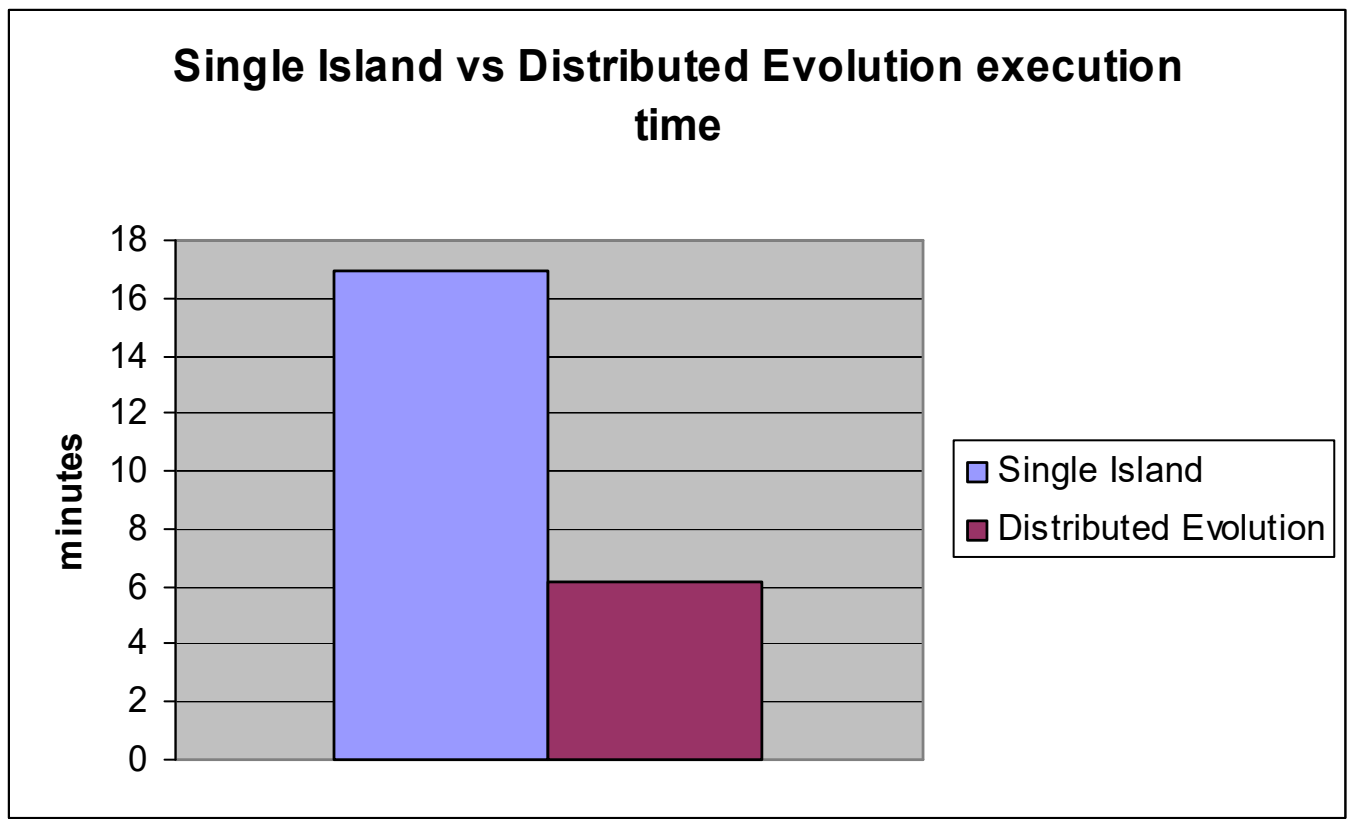

**Figure 5. Single Island vs. Distributed Evolution Execution Time**

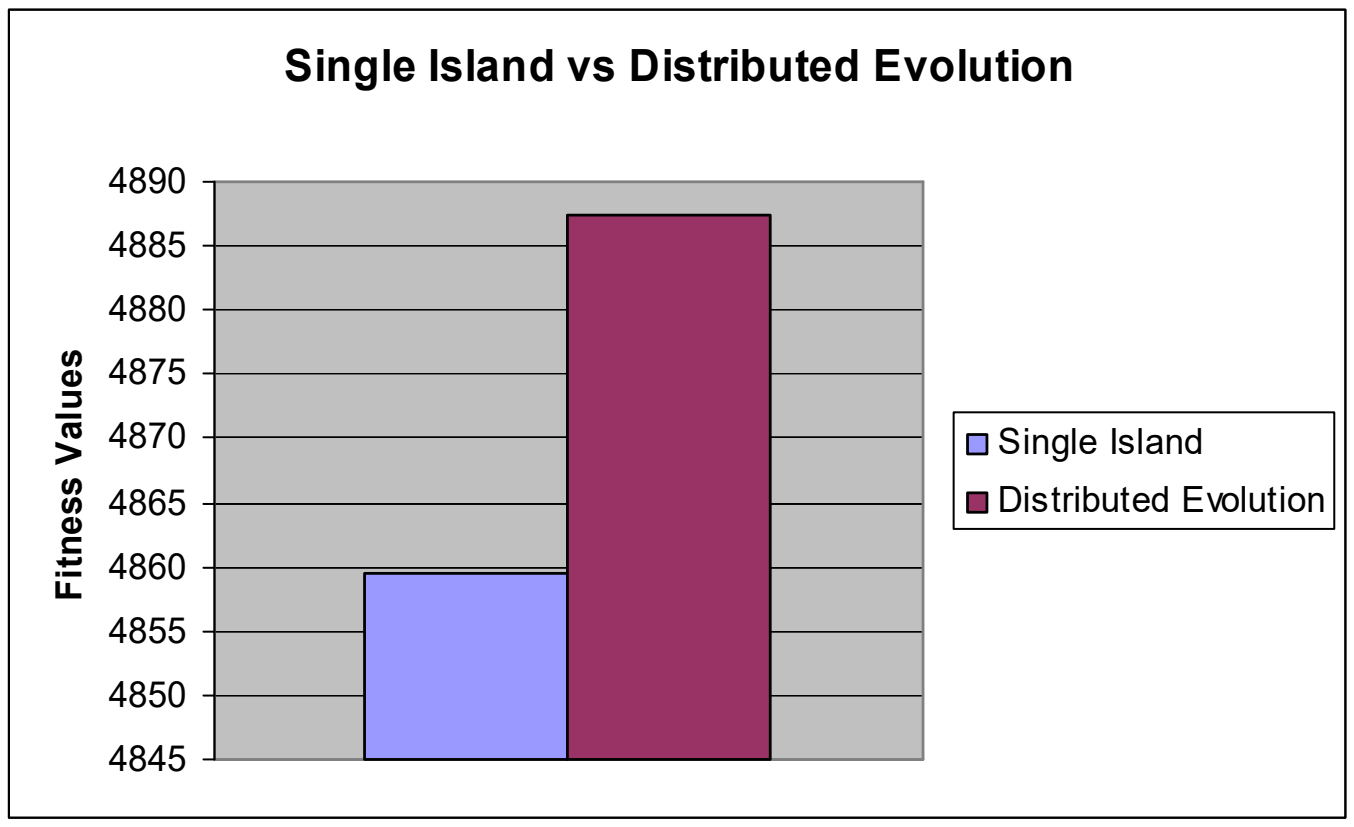

**Figure 6. Single Island vs. Distributed Evolution Fitness Values**

As seen in charts from figures 5 and 6, by running the same number of evaluations on a single island and running them distributed over a set of islands (four in this case) we can see that the gain in fitness is superficial but the loss in time is three times bigger. The low increase on the fitness obtained can be backed up by the lack of diversity existent in a single population.

## CONCLUSIONS

The conclusions we were able to obtain from the tests performed, where as expected that with the use of migrations over a set of parallel genetic algorithms, we can improve the diversity of the chromosomes in each island, leading to better results, and avoiding local maximums (in our case local minimums due to the problem specification to minimize the search). As a result of this, the convergence is delayed and the probability of obtaining better results is increased.

Increasing the number of migrations on the same search, we can see that in the limit we will be approaching the single island model; since migrations occur so frequently that the diversity of chromosomes is the same in all islands.

We can also conclude that the best timing in which a migration should occur is when the deviation approaches zero meaning that we have already exhaustively explored that local maximum. In this case a migration would bring the necessary heterogeneity through new and different chromosomes with fresh genotypes. Allowing the search to overcome the local maximum, and proceed with its search of a better solution.

From the second set of tests we can conclude that with the parallel solution without migrations we can obtain the same results as can be obtained running the same number of evaluations on a single island, but much faster.

## ACKNOWLEDGMENTS

We would like to thanks Eng. João Paulo Caldeira for his availability and help every time we needed it.